\documentclass[runningheads]{llncs}
\usepackage{graphicx}
\usepackage{caption}
\usepackage{subcaption}
\usepackage{booktabs}
\usepackage{wrapfig}

\begin{document}
\title{Evaluation of Human-Understandability of Global Model Explanations using Decision Tree}
\titlerunning{Human-Understandability of Global Model Explanations}
%
\author{Adarsa Sivaprasad\inst{1}\orcidID{0000-0002-4460-1671} \and Ehud Reiter\inst{1}\orcidID{0000-0002-7548-9504} \and Nava Tintarev\inst{2}\orcidID{0000-0003-1663- 
        1627} \and Nir Oren\inst{1}\orcidID{0000-0002-4854-9014}
 }
\authorrunning{A. Sivaprasad et al.}
%
\institute{Department of Computing Science, University of Aberdeen, Aberdeen, UK \email{\{a.sivaprasad.22,e.reiter,n.Orin\}@abdn.ac.uk}
\and Maastricht University, Maastricht, Netherlands \email{n.tintarev@maastrichtuniversity.nl}}

\maketitle 

\begin{abstract}
In explainable artificial intelligence (XAI) research, the predominant focus has been on interpreting models for experts and practitioners. Model agnostic and local explanation approaches are deemed interpretable and sufficient in many applications. However, in domains like healthcare, where end users are patients without AI or domain expertise, there is an urgent need for model explanations that are more comprehensible and instil trust in the model's operations. We hypothesise that generating model explanations that are narrative, patient-specific and \emph{global}(holistic of the model) would enable better understandability and enable decision-making. We test this using a decision tree model to generate both local and global explanations for patients identified as having a high risk of coronary heart disease. These explanations are presented to non-expert users. We find a strong individual preference for a specific type of explanation.  The majority of participants prefer global explanations, while a smaller group prefers local explanations. A task based evaluation of mental models of these participants provide valuable feedback to enhance narrative global explanations. This, in turn, guides the design of health informatics systems that are both trustworthy and actionable.

\end{abstract}
\section{Introduction}
The field of explainable artificial intelligence (XAI) has witnessed significant advancements, primarily focusing on the interpretability of models. However, the interpretability of an AI model for developers does not seamlessly translate into end-user interpretability~\cite{Biran2017ExplanationAJ}. Even inherently interpretable models like decision trees(DT) and decision lists are challenging to use in applications due to the complexity and scale of data. Hence popular explanation techniques interpret black box models by considering an individual input and corresponding prediction - \emph{local explanations}. Model-agnostic explanations such as Shapley values and Local Interpretable Model-Agnostic Explanations (LIME) offer insights into the features contributing to an individual prediction, revealing the importance of specific characteristics in decision-making. Nevertheless, they do not capture the complete model functioning, comprehensive utilization of data, and, most importantly, the interactions among features. They lack the ability to facilitate generalization or provide a complete mental model of the system's workings.

\begin{figure}
    \includegraphics[width=1\linewidth]{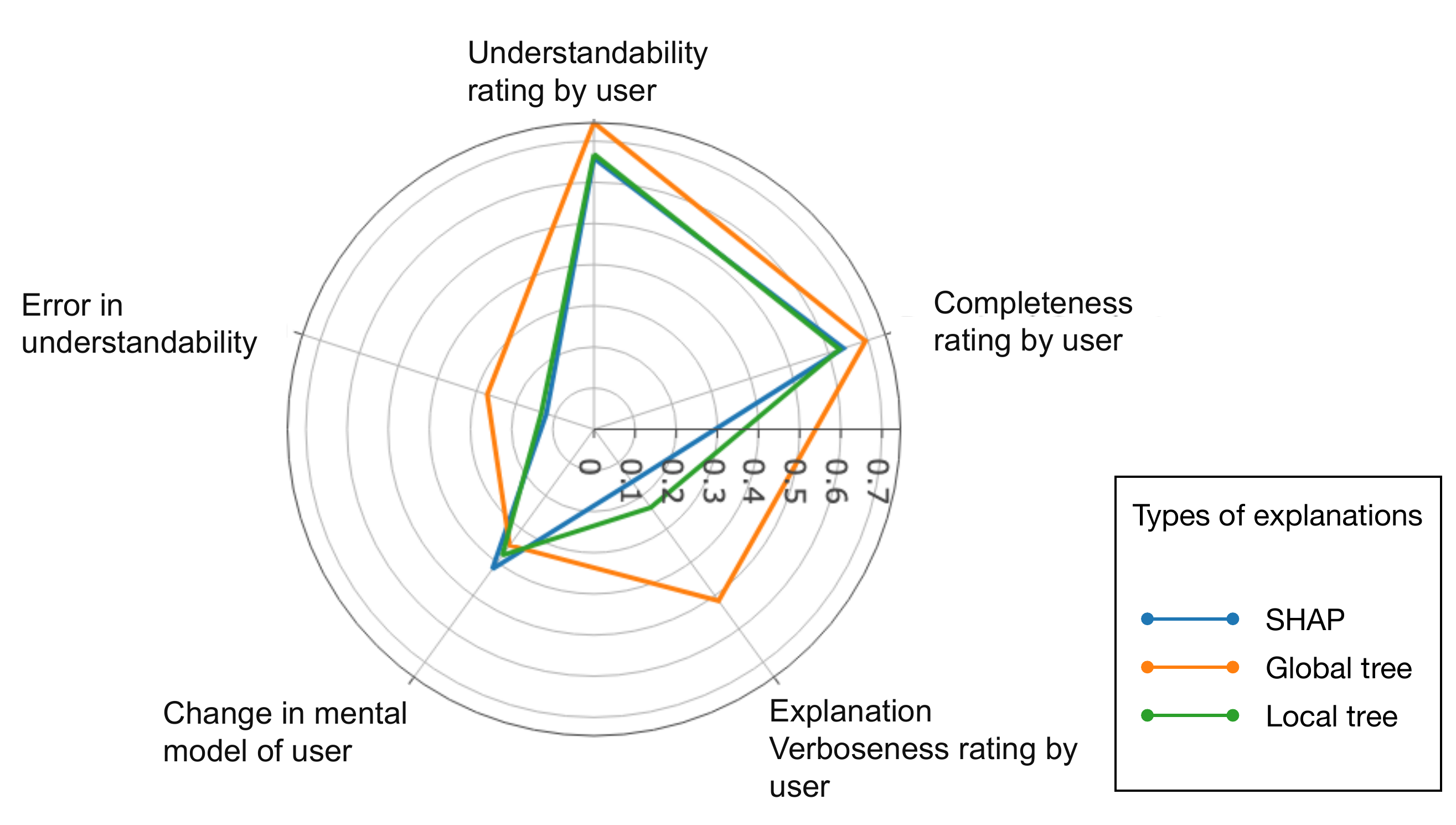}
    \caption{A comparison of Local SHAP, Local and Global tree explanation of CHD risk prediction using decision tree model. Different evaluation parameters are computed based on end-user feedback of the explanation.}
    \label{fig:preview}
\end{figure}

In critical domains such as healthcare and financial predictions, the interpretability of AI models by end-users holds significant importance. The understandability of the underlying AI model and the trust in its predictions can have life-altering implications for stakeholders. Enabling user intervention and action to modify predicted outcomes require explanations that address the \emph{How} and \emph{Why} questions, as well as convey causal relationships \cite{MILLER20191,10.5555/3238230}. Achieving this necessitates an overall comprehension of the model. Further, the explanation should not only align the user's mental model with the AI system's model but also be perceived as understandable and trustworthy. We propose that a global model explanation hold greater potential for providing understandability and building trust compared to local model explanations. This study is a preliminary step towards testing this.

What qualifies as a global explanation and what methodologies would provide an overall understandability is relatively less researched. The comparison between global model explanations and local explanations for end users, along with various presentation aspects such as narrative and visualization, bears significance when building explanation-centric applications. This study delves into the understandability of local and global explanations, specifically in the context of a coronary heart prediction model. We address the following research question: 
\begin{enumerate}
\item For non-expert users, do global explanations provide a better understanding of the AI's reasoning in comparison to (only) local explanations?
\item  As the complexity of the explanation increases is there a difference in understandability and user preference for local and global explanations?
\end{enumerate}

We use decision tree (DT) models which are interpretable by design, and construct local and global explanations with varying levels of complexity. We gauge the perceived understandability of these models and evaluate their effectiveness based on predefined tasks. We also measure the changes in users' mental models following exposure to the explanations. Figure \ref{fig:preview} shows different evaluation parameters. The experiment identifies preferences in explanation types among different participant groups. It is found that while complexity does not have a significant effect on perceived understandability and completeness of explanation, errors in understanding increase with complexity. The obtained results offer valuable insights for designing narrative explanations for end-users and highlight the majority of participant preference for global explanations in healthcare risk models. 

\section{Related Work}

In healthcare, a risk score is a quantifiable measure to predict aspects of a patient’s care such as morbidity, the chance of response to treatment, cost of hospitalisation etc. Risk scoring is utilised for its predictive capabilities and in managing healthcare at scale. A predicted risk score is representative of the probability of an adverse outcome and the magnitude of its consequence. Article 22 of the General Data Protection Regulation (GDPR) mandates human involvement in automated decision-making and in turn understandability of a risk prediction model. Hence the use of risk scores requires the effective communication of these scores to all stakeholders - doctors, patients, hospital management, health regulators, insurance providers etc. With statistical and black-box AI models used in risk score computations, this is an added responsibility of the AI model developer to ensure the interpretability of these systems to all stakeholders. 

Current regulations such as model fact tables~\cite{Sendak2020PresentingML} are useful for clinicians and approaches of local model interpretation~\cite {lundberg_unified_2017,ribeiro-etal-2016-trust} to model developers. For a non-expert end-user who has limited domain knowledge and who is not trained to understand a fact table, these approaches will not explain a recommendation given to them. Further, explaining a risk prediction model to the end user should address the perceived risk from numeric values and previous knowledge of the user, any preferences and biases. In other words, the explanation presentation should address socio-linguistic aspects \cite{MILLER20191} involved.

Researchers have recognized that a good explanation should aim to align the user's mental model with the mental model of the system, promoting faithful comprehension and reducing cognitive dissonance~\cite{MILLER20191}. Achieving such effectiveness is very context-dependent~\cite{surveyXAI}. However, aspects of explanation presentation generalise across a broad spectrum of applications. The significance of narrative-style explanations is emphasised by~\cite{reiter-2019-natural} while~\cite{doi:10.1146/annurev-statistics-010814-020148}, highlights the effectiveness of a combined visual and narrative explanation. Recent studies have evaluated existing systems in use ~\cite{doi:10.1146/annurev.publhealth.28.021406.144123,MARKUS2021103655} and calls for focus on the design choices for explanation presentation in health informatics. Further,  with tools available in the public domain such as QRisk\footnote{\url{https://qrisk.org/index.php}} from National Health Service (NHS), evaluating the impact and actionability of explanation approaches in use would enable improving them and ensure their safe usage. 

Before looking into evaluating black-box models, it would be worthwhile to explore what constitutes a good explanation in interpretable models such as DTs, decision lists~\cite{decisioLists} etc.  DT algorithms are methods of approximating a discrete-valued target by recursively splitting the data into smaller subsets based on the features that are most \emph{informative} for predicting the target. DTs can be interpreted as a tree or as a set of if-else rules which is a useful representation for human understanding. The most successful DT models like Classification and Regression Trees (CART)~\cite{brei} and C4.5~\cite{quinlan1993c45} are greedy search algorithms. Finding DTs by optimising for say a fixed size,  is NP-hard, with no polynomial-time approximation~\cite{journals/ipl/HyafilR76}. Modern algorithms have attempted this by enforcing constraints such as the independence of variables~\cite{KlivansDecisionList} or using all-purpose optimization toolboxes ~\cite{Bertsimas2017,Blanquero_2021,Verwer_Zhang_2019}.

In~\cite{10.1145/2939672.2939874} authors attempt the optimisation of the algorithm for model interpretability to derive decision lists. The reduced size of the rules opens up the option of interpreting the decisions in their entirety and not in the context of a specific input/output alone - a global explanation. The authors highlight the influence of complexity on the understandability of end-users. However, decision list algorithms still do not scale well for larger datasets.
Optimal Sparse Decision Trees (OSDT)~\cite{NEURIPS2019_ac52c626} and later improved with Generalized and Scalable Optimal Sparse Decision Trees (GOSDT)~\cite{GOSDT} algorithms produce optimal decision trees over a variety of objectives including F-score, AUC, and partial area under the ROC convex hull. GOSDT generates trees with a smaller number of nodes while maintaining accuracy on par with state-of-art models. 

On explaining DTs for end-users, current studies have investigated local explanations using approaches such as counterfactuals~\cite{DBLP:journals/corr/abs-1711-00399}, the integration of contextual information and identified narrative style textual explanations\cite{MARUF2023101483}. All these attempts to answer the \emph{why} questions based on a few input features and specific to a particular input. Extending these insights to global explanations should help better understanding of the model by end-users and allow generalisation of the interpretations, driving actionability. 

\section{Experiment Design}

Our main research question is to determine what type of explanation are most relevant for non-expert end-users to be able to understand underlying risk model. We evaluate a local and global explanation by measuring user's perceived understanding and completeness. We also measure whether the user's mental model had changed after reading an explanation.

\subsection{Dataset and Modeling}
For the experiment, we used the Busselton dataset~\cite{knuiman1998multivariate}, which consists of 2874 patient records and information about whether they developed coronary heart disease (CHD) within ten years of the initial data collection. This study is similar to the data collected by NHS to develop  QRISK3~\cite{Hippisley-Coxj2099}. Computing a risk score demands that we also explain the risk score,  data used, probability measures of the scoring algorithm in addition to model prediction. We limit the scope of this study to only explaining the model prediction and use the CHD observation from the dataset as target variable for prediction. Using GOSDT~\cite{GOSDT} algorithm, we fit the data to obtain decision tress. GOSDT handles both categorical and continuous variables. While the optimum model may have multiple closeby splits for numeric values, such splits can reduce the readability of the tree. Hence we preprocess the data by converting most of the features into categorical variables. We follow the categories as mandated by National Health Service(NHS). The data is pre-processed as described in Appendix \ref{appendix:ConstDT}, with 2669 records and 11 features.

The GOSDT algorithm generated a comprehensive decision tree for the dataset, comprising 19 leaf nodes at a depth of 5, achieving an accuracy of 90.9\% (Figure \ref{fig:DTall} in Appendix \ref{appendix:ConstDT}). For the purpose of human evaluation and comparison of local and global explanations, it was necessary to have multiple DTs with comparable structures. Hence, we created subsets of the data by varying the ranges and combinations of \emph{Age} and \emph{Gender}. By working with reduced data points, the size of the constructed trees was significantly reduced. To ensure larger trees for evaluation purposes, we enforced a consistent depth of 4. Ultimately, we selected four trees for the evaluation task as shown in Table \ref{tab:TreeDescription}.

As mentioned in~\cite{DBLP:journals/corr/abs-1802-00682}, a higher complexity of explanation rules in clinical setting leads to longer response times and decreased satisfaction with the explanations for end-user. The authors refer to unique variables within the rules as cognitive chunks, which contribute to complexity in understanding. In our experiment, global explanations naturally contain more cognitive chunks. To prevent bias in the results, we incorporated two levels of difficulty for each explanation type.  The easy level consisted of trees with similar structures, both local and global, featuring 5 nodes and decision paths of equal length with an identical number of cognitive chunks. For ease of understanding, we henceforth refer to a particular combination of explanation type and difficulty level as a specific scenario, namely - local-easy, global-easy, local-hard, and global-hard. A local-SHAP explanation was generated utilizing the same tree as the local-easy scenario. We use kernel SHAP~\cite{lundberg_unified_2017} to obtain feature importance for the local-easy tree for specific patient input. The SHAP explanation is treated as a baseline for evaluation.

The hard scenario for both explanation types, consist of larger trees of similar structures. The tree had 8 nodes for local-hard scenario and 9 nodes in case of global-hard scenario. For global explanations, the explanation presentation involves more cognitive chunks, potentially introducing bias by making the global-hard scenario challenging to comprehend. Nevertheless, we proceeded with evaluating this scenario in our experiment.

Another factor to consider when generating explanation is the possible contradiction between model explanation and general assumptions. For instance, a node  \emph{BMI = Normal} appearing in decision rules for low CHD risk is expected but not in those for high risk. Communicating this contradiction in explanation would be important in its understandability. We also include this in our experiment. Explanation scenarios categorized as hard involved contradictory explanations, which could prove more challenging for comprehension. We addressed these cases using semifactual \cite{MORENORIOS2008197} explanations, employing phrase \emph{even-if}. We assess the impact of such risk narrations on understandability. Table~\ref{tab:TreeDescription} provides a summary of the four trees used for explanation generation.

\begin{table}
\caption{{Description of DTs and type of explanation generated.}}
\label{tab:TreeDescription}
\centering
\begin{tabular}{p{2cm}p{2cm}p{2cm}p{2cm}p{3.5cm}}
\toprule
\textbf{Age} & \textbf{Gender}& \textbf{Leaf count} & \textbf{Accuracy}& \textbf{Explanation Type} \\
\midrule
\midrule
70 - 79 & Female & 6 & 78.4 & Local Easy\\
\hline
60 - 84 & Female & 6 & 82.5 & Global Easy \\
\hline
60 - 70 & Male & 9 & 77.3 & Local Hard \\
\hline
65 - 70 & male & 10 & 85.4 & Global Hard \\
\bottomrule
\end{tabular}
\end{table}

\subsection{Generation of Explanation}
\begin{figure}
     \centering
     \begin{subfigure}[b]{\textwidth}
         \centering
         \includegraphics[width=\textwidth]{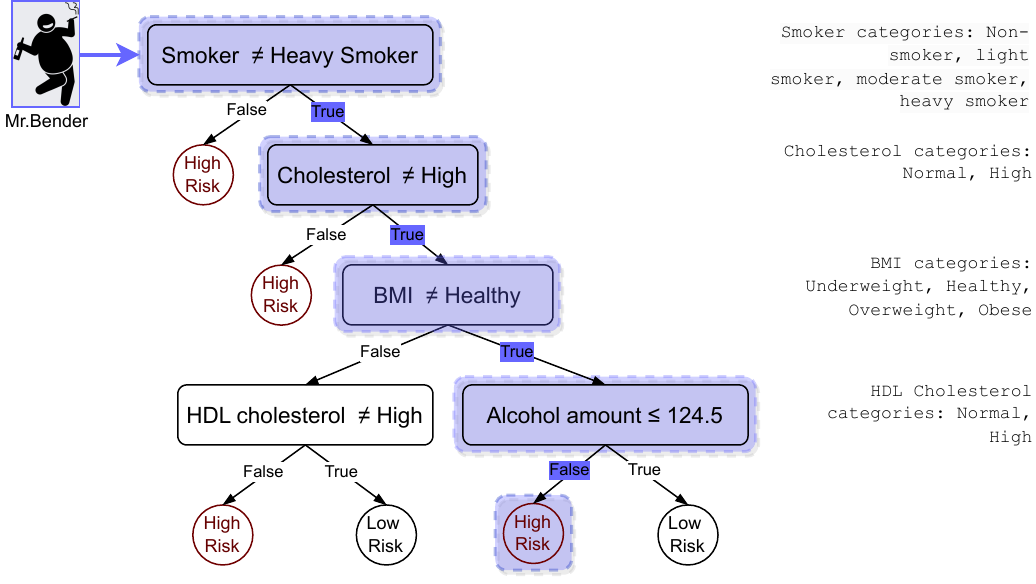}
         \caption{The decision path followed along a given DT for a particular patient Mr Bender. The tree is learned from different categorical features of a patient dataset  and the black square boxes represent decision nodes learned by the model. On the right, all the possible values of each feature are listed (except Alchohol amount which is numeric). This tree has 6 leaf nodes each a possible decision of \emph{high} or \emph{low risk}. For a given input corresponding to Mr Bender, the model predicts \emph{high risk} following the decision path highlighted in Blue.    }
         \label{fig:XGen1}
     \end{subfigure} \par\bigskip
     \hfill
     \begin{subfigure}[b]{\textwidth}
         \centering
         \includegraphics[width=\textwidth]{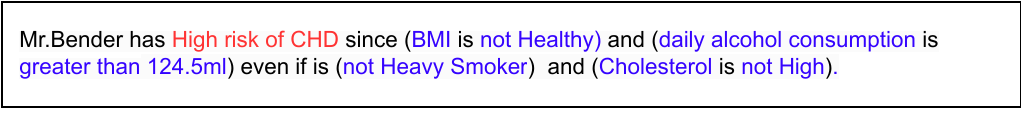}
         \caption{A local explanation of the decision in \ref{fig:XGen1}.}
         \label{fig:XGen2}
     \end{subfigure} \par\bigskip
     \hfill
     \begin{subfigure}[b]{\textwidth}
         \centering
         \includegraphics[width=\textwidth]{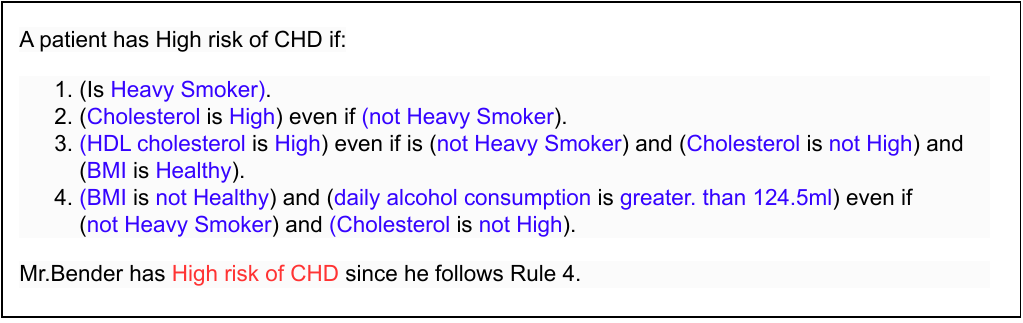}
         \caption{A global explanation of the DT and the decision in \ref{fig:XGen1}.}
         \label{fig:XGen3}
     \end{subfigure}
        \caption{An example of local and global narrative explanation of a DT. Note that this is one way of generating a global tree explanation (Appendix\ref{appendix:genExpNarration}). Listing all the nodes or stating all possible categorical values of features are design choices that will  affect understandability.}
        \label{fig:three graphs}
\end{figure}

For a given CHD prediction model and a corresponding patient input, the local explanation is a set of necessary conditions and predicted decisions of high/low risk. For the decision tree model in Figure~\ref{fig:XGen1}, given particular patient info as input, the decision rule that is triggered to predict high risk is highlighted in blue. The path followed for the decision can be represented textually as shown in Figure~\ref{fig:XGen2}. This is one possible representation. A more natural language expression of the rule is treated as a local explanation for the experiment. The language generation is rule-based. Details of the generation algorithm and an example of the evaluated explanation are given in Appendix~\ref{appendix:genExpNarration}.

The global tree explanation is a list of all the decision rules of the tree. For a particular patient, a combination of the global explanation and the specific rule triggered corresponding to the given patient input is treated as the global prediction explanation. Once again, this is a choice we make for this experiment. A list of all decision nodes similar to feature importance in SHAP could also be a possible global tree explanation. For the patient in Figure~\ref{fig:XGen1}, the corresponding global explanation is shown in Figure~\ref{fig:XGen3}. As the tree size becomes large, the number of rules and the number of features in each rule increase. This means the explanation size and the cognitive chunks in the explanation increase. The best way to frame natural language explanations, for these different cases, is a separate research problem that we do not address here. Further, we restrict the rules in global explanation to those corresponding to a single risk category -  high risk. Since the particular case involves only two categories, this still provides coverage to possible predictions while keeping the explanation less verbose. The narration generation involves the same algorithm as in the case of local explanation.

In addition to the model accuracy, note that each leaf node has a probability and confidence associated with that particular decision. For a particular node, the probability is the ratio of training data points that fits the criteria of that node to the number of data points in its previous node. A low probability node denotes that, the particular decision was rare based on the training data. The statistical significance of this prediction denotes its confidence. Both these measures are used for generating decision narration. Appendix \ref{appendix:genExpNarration} shows examples of the usage. To express the probabilities, we use verbal mapping proposed by ~\cite{doi:10.1146/annurev-statistics-010814-020148}. An additional usage of \emph{possibly} is introduced to accommodate cases involving low confidence and high probability.   

The SHAP explanation does not have associated confidence. We filter features with SHAP score greater that 0 and present them as bulleted points in descending order of importance.

\subsection{Evaluation}
For evaluation, a within-subject survey is conducted with participants recruited on Prolific platform. We conducted a pilot study among peers and the feedback was used to improve the readability of the explanations and assess the time taken for the tasks. 

The survey involves 5 patient scenarios namely local-SHAP, local-easy, local-hard, global-easy and global-hard. Each scenario consists of 2 pages. On the first page, the participant is provided with information about a patient. This consists of their features: age, gender, height, weight, BMI, blood pressure, different cholesterol values, smoking, and drinking habit. They are asked to enter the assumptions on what patient features may contribute to the AI model’s prediction. This captures the mental model of the participant regarding CHD. Appendix \ref{appendix:userSurvey} shows examples of the pages used in the survey.

On the next page, participants are presented with the same patient, the risk of CHD (high or low) as predicted by the AI system along with an explanation. They are asked to enter feature importance once again based on their understanding of the explanation. They are also asked to rate the explanation on three parameters: completeness, understandability, and verboseness,  using a 5-level Likert scale. Text feedback on each explanation and overall feedback at the end of the survey is collected. 

\begin{table}[t]
\caption{Evaluation criteria for comparison of different explanation types.}
\label{tab:evalParam1}
\centering
\begin{tabular}{p{5cm}p{9cm} }
\toprule
Measure & Definition\\
\midrule
\midrule
\emph{Completeness rating (CR)} & User rating for the prompt: This explanation helps me completely understand why the AI system made the prediction \\
\midrule
\emph{Understandability rating (UR)} & User rating for the prompt: Based on the explanation I understand how the model would behave for another patient  \\
\midrule
\emph{Verboseness rating (VR)} & User rating for the prompt: This explanation is long and uses more words than required \\
\midrule
\emph{Change in mental model (CMM)} & Difference in perceived feature importance before and after viewing model explanation\\
\midrule
\emph{Error in Understanding (EU)} & Difference between model feature importance and perceived feature importance after viewing explanation \\
\bottomrule
\end{tabular}

\end{table}

The evaluation of each explanation has 3 parameters from a Likert rating based on participant perceptions. In addition, based on the task of choosing feature importance we compute two additional parameters: change in mental model and correctness of understanding. Change in mental model is defined as the updation of perceived feature importance before and after explanation. Let \(U = (u_1, u_2, ..., u_N)\) where \(u_i\in{\{0,1\}}, 1\le i\le N \) be the selected feature importance before explanation where N is the total number of features. Let \( V = (v_1, v_2, ..., v_N) \) where \(v_i \in {\{0,1\}}, 1\le i \le N \) be the selected feature importance after explanation. \emph{Change in mental model} is computed as \[D_m = \frac{d(U,V)}{N} \] where \(d\) is the \emph{Hamming distance} between \emph{U} and \emph{V}. 

For each explanation, based on the features that are shown in the narration, we also know the \emph{correct} feature importance. In the case of SHAP, these are the features with a SHAP score greater than 0. For local explanations, these are the features in the decision path, and for global explanations, it is all the features in the tree. If the correct feature importance \( C = (c_1, c_2, ..., c_N)\) where \(c_i \in {\{0,1\}}, 1\le i \le N \), we compute the \emph{error in understanding} w.r.t to the system mental model as \[D_c = \frac{d(V,C)}{N} \]. Since for each feature, the participant selects a yes/no for importance, these measures do not capture the relative importance among features. Table \ref{tab:evalParam1} summarises all the evaluation parameters.

\section{Results and Discussion}
Fifty participants were recruited from the Prolific platform for the experiment, ensuring a balanced gender profile. All participants were presented with five patient-explanation scenarios and were requested to evaluate each of them. The survey took an average of 26 minutes to complete, and participants received a compensation of £6 each, as per the minimum pay requirement. However, one participant was excluded from the analysis due to indications of low-effort responses, spending less than 1 minute on multiple scenarios. The demographic details of the selected participants are summarized in Table \ref{tab:particiantDemography}.
\vspace{-1.5em}
\begin{table}
\caption{Demographic distribution of survey participants.}
\label{tab:particiantDemography}
\centering
\begin{tabular}{p{2.5cm}|p{9.5cm}}
\toprule
\textbf{Feature} & \textbf{Category: Proportion} \\
\midrule
\midrule
Age & 18-30 : 81.63\%, 30-40 : 16.33\%, 40-65 : 2.04\% \\
\hline
Gender & Male : 51.02\% , Female : 48.98\% \\
\hline
First language & English:38.8\%, Others:61.2\%  \\
\bottomrule
\end{tabular}

\end{table}
Based on the responses, we computed the evaluation parameters mentioned in the previous section. The Likert scale ratings for \emph{Completeness}, \emph{Understandability}, and \emph{Verboseness} are assigned values from 0 to 1, 0 corresponding to 'Strongly Disagree' and 1 to 'Strongly Agree'. We also calculate, \emph{Change in the mental model} and \emph{Error in understanding} from the selection of feature importance. The calculated scores are also normalised to range from 0 to 1. The mean values across all participants are presented in Table \ref{tab:scenarioMean}.
\vspace{-1.5em}
\begin{table}
\caption{Evaluation parameters across all the scenarios. Maximum is highlighted in bold and minimum in italics. CR -  Completeness rating, UR - Understandability rating, VR - Verboseness rating, CMM - Change in mental model, EU - Error in Understanding. }
\label{tab:scenarioMean}
\centering
\begin{tabular}{ p{1.5cm}p{2.4cm}p{2.4cm}p{2.4cm}p{2.4cm}p{2.4cm}  }
\toprule
&Local SHAP & Local Easy & Local Hard& Global Easy & Global Hard \\
\midrule
\midrule
\textbf{CR} & 0.64 & 0.69 & \emph{0.63} & 0.68 & \textbf{0.69} \\
\midrule
\textbf{UR} & \emph{0.66} & 0.71 & 0.67 & 0.72 & \textbf{0.74}\\
\midrule
\textbf{VR} & \emph{0.16} &0.26 & 0.23 & \textbf{0.56} & 0.52\\
\midrule
\textbf{CMM} & \textbf{0.42} & \emph{0.28} & 0.38 & 0.34 & 0.35\\
\midrule
\textbf{EU} & 0.12 & \emph{0.07} & 0.13 & 0.19 & \textbf{0.30}\\
\bottomrule
\end{tabular}
\end{table}
\vspace{-1em}

While local-easy scenario has the lowest error in understandability(EU), participants rated all the models comparably in terms of Understandability (UR) and Completeness (CR). The Change in the Mental Model (CMM) exhibited uniformity across all types of explanations, except for local-SHAP and local-easy. 
To assess the significance of these results, we performed the Wilcoxon test, for all combinations of explanation types. Since multiple comparisons are performed, we apply Bonferroni Correction on p-value and a threshold of 0.01 is chosen. In comparing local and global explanations, local-SHAP is excluded and the ratings for both levels of difficulty in each case are averaged. The results are shown in Table \ref{tab:significanceTest}. The observations that hold for a stricter threshold of 0.001 are highlighted with $*$.

\begin{table}[htbp]
\caption{Significance of difference between types of explanation. CR -  Completeness rating, UR - Understandability rating, VR - Verboseness rating, CMM - Change in mental model, EU - Error in Understanding. The values which are significant (Bonferroni Corrected p-value threshold of 0.01) are highlighted in \textbf{bold}. P-value $\leq{0.001}$ are highlighted with *.}
\label{tab:significanceTest}
\centering
\begin{tabular}{p{5cm}p{1.7cm}p{1.7cm}p{1.7cm}p{1.7cm}p{1.7cm}}

\toprule
& CR & UR & VR & CMM & EU \\
\midrule
\midrule
Local vs Global & 0.42 & 0.44 & $\textbf{0.00}^*$ & 0.53 & $\textbf{0.00}^*$ \\
\midrule
Local Easy vs Global Easy & 0.84 & 0.85 & $\textbf{0.00}^*$ & 0.05 & $\textbf{0.00}^*$ \\
\midrule
Local Hard vs Global Hard & 0.35 & 0.42 & $\textbf{0.00}^*$ & 0.36 & $\textbf{0.00}^*$ \\
\midrule
Local Easy vs Local Hard & 0.38 & 0.24 & 0.76 & $\textbf{0.00}^*$ & $\textbf{0.00}^*$ \\
\midrule
Global Easy vs Global Hard & 0.50 & 0.53 & 0.56 & 0.43 & $\textbf{0.00}^*$ \\
\midrule
Local SHAP vs Local Hard & 0.63 & 0.76 & 0.10 & 0.23 & 0.42 \\
\midrule
Local SHAP vs Local Easy & 0.18 & 0.28 &0.03 & $\textbf{0.00}^*$ & 0.11 \\
\midrule
Local SHAP vs Global Hard & 0.02 & 0.30 & $\textbf{0.00}^*$ & 0.09 & $\textbf{0.00}^*$ \\
\midrule
Local SHAP vs Global Easy & 0.16 & 0.28 & $\textbf{0.00}^*$ & \textbf{0.01} & 0.02 \\
\bottomrule
\end{tabular}
\end{table}

Global explanations resulted in a lower average understandability based on the feature selection(EU) and it was observed that harder scenarios resulted in higher errors for both local and global explanations. For each type of explanation, the patient features wrongly selected was investigated (Table \ref{tab:errorType1}, \ref{tab:errorType2}). Incorrect feature selection related to \emph{cholesterol}  caused the majority of errors. Participants chose the wrong cholesterol-related feature, possibly due to a lack of attention or limited understanding of medical terminology. Improving the presentation of explanations and providing more contextual information could potentially address this issue. Importantly, when presented with semifactual explanations of hard scenarios both local and global explanations led to almost half or more participants excluding the corresponding feature. This clearly points to the ambiguity of such narration.

\begin{figure}
    \centering
    \includegraphics[width=1\linewidth]{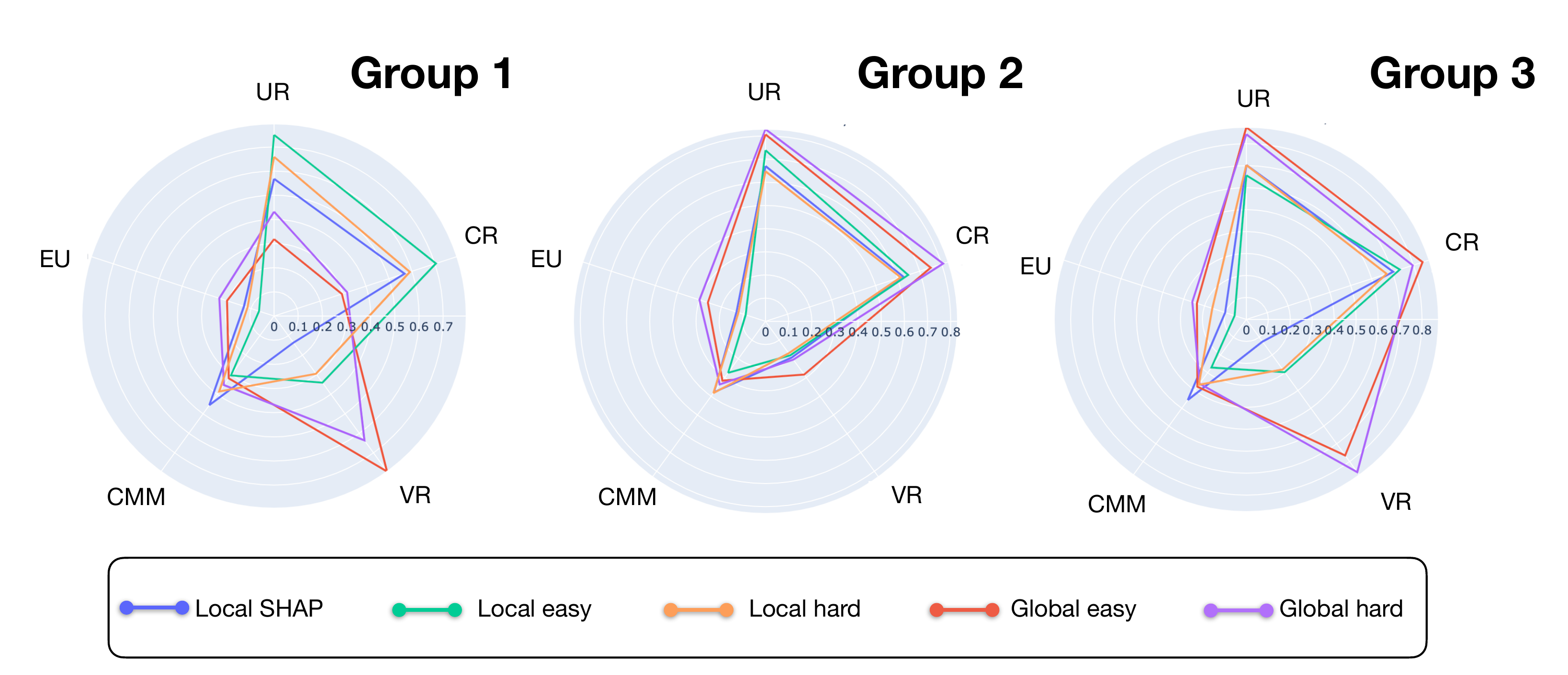}
    \caption{Average rating for different explanation type across the participant groups}\label{fig:clusterDist}
\end{figure}

The error analysis does not explain the contradiction between the understandability ratings and the correctness of feature selection. Interestingly, a considerable number of participants expressed a preference for longer, global explanations, even if they did not fully comprehend them. Significant rating of global explanations as more verbose adds to this contradiction. To delve deeper into this phenomenon, participant clustering was performed based on the ratings and computed scores. Using the k-means algorithm, three distinct groups of participants were identified and manually validated. Figure \ref{fig:clusterDist} displays the average rating across different parameters for each group.

\begin{itemize}
   \item Group 1: Strongly prefer and understand local explanations. The cluster consists of 11 participants who rate patient-specific local tree explanations highest on completeness and understandability.
   \item Group 2: Majority group that rates global explanation as most understandable: This cluster consist of 22 people who has the least significance in preference between global, local explanation or difference based on the difficulty level. They rate Global explanation highest on completeness and understandability
   \item Group 3: This cluster consist of 16 people who strongly prefer global explanations but critical about the narration. This cluster is more detail oriented and rates global explanations as more understandable and complete.This group was critical on the narration and presentation of explanation in the feedback form. The average error in feature selection for global explanation for this group, is lower than Group 2.
 \end{itemize}
It is evident that within the clusters, the ratings on each parameters  has significant preferential pattern between each type of explanation. Group 1, 3 has strong polarity on the preferences and their rating tend to \emph{Strongly agree} or \emph{Strongly disagree}. Both these Groups identify Global explanations as verbose. This shows that, in healthcare setting, the effectiveness of an explanation to an end-user, is very dependent on their individual preference.

\subsection{Local vs. Global}

While there is no significant difference between local and global explanations overall,  strong differences emerge at the Group level. Group 1 rates the local explanation as complete, while both Groups 2, and 3 favour the global explanation for completeness. Similar preferences are observed in participants' perception of understandability within each group. When a stricter p-value threshold of 0.001 is applied, the significance of the difference in user rating for understandability and correctness holds only in Group 1.  The results of the Wilcoxon test for combinations of explanation types within Groups are given in Appendix \ref{appendix:resultsXTnded}. 
\begin{itemize}
\item \emph{The results indicate that certain people strongly prefer specific type of explanation. This preference does not necessarily translate to understandability.}
\item \emph{In all groups, a higher error in feature selection is observed for global explanations, mainly due to the semifactual explanation and wrongly interpreting features related to Cholesterol}
\end{itemize}

 Among participants belonging to Group 2, the factors driving their preference for global explanations remain unclear. Demographics data examination (Table \ref{tab:groupDemo}) offers no apparent patterns, leading us to propose the influence of unique cognitive styles within the groups. Further investigations are warranted to unveil the underlying reasons for these preferences and errors. While users may perceive explanations as understandable, it is vital to recognize that this perception may not necessarily translate into accurate decision-making. The lack of significant changes in mental models substantiate this, indicating the need for continued exploration to optimize explanation presentations for healthcare AI models.
\vspace{-1.5em}
\begin{table}
\caption{Demographic distribution of participants within each group. All the features are not available for all participants. Missing data are excluded in the counts. }
\label{tab:groupDemo}
\centering
\begin{tabular}{p{5.5 cm}|p{2cm}p{2cm}p{2cm}}
\toprule
&\textbf{Group1} & \textbf{Group2} & \textbf{Group3} \\
\midrule
Number of participants &11 &22 &16 \\
\hline
Male to female ratio & 4:7 & 9:13 &12:4 \\
\hline
Count of full time employed &2& 8&5\\
\hline
Student to non-student ratio & 8:2 & 10:9&8:7 \\
\hline
Number of native english speakers &4 &11&4 \\
\hline
Ethnicity, white to black ratio &9:2 &11:10&11:3 \\
\bottomrule
\end{tabular}
\end{table}
\vspace{-1.5em}
\subsection{Tree Explanation vs. SHAP}
The overall ratings of SHAP explanations are comparable to those of local-hard explanations but lower than those of local-easy explanations generated from the same underlying decision tree. This suggests that the comprehensibility and interpretability of SHAP explanations are slightly lower than those of the local-easy explanations. However, this may be attributed to the presentation bias, as all participants were exposed to the SHAP explanation first. 
It is noted that the presentation style of SHAP explanations, using bulleted points, is generally considered less verbose even though it does not impact the error in understandability or perceived understandability and completeness. Hence the simpler readability of the SHAP explanation is not seen to have impacted its overall understandability.

\subsection{Easy vs. Hard}

The ratings provided by the participants on the Likert scale did not reveal any significant distinction between the explanation scenarios characterized as easy and hard. However, an examination of the impact of difficulty levels on the error in feature selection uncovered significant results. Hard scenarios, whether global or local explanations, exhibited significantly higher error rates, even within participant groups.

\begin{itemize}
\item \emph{The explanation understanding is strongly dependent on the complexity of the feature interaction being explained.}
\end{itemize}

When participants encountered explanations that deviated from their preexisting notions of feature dependence, it introduced confusion, becoming a major contributor to error in hard scenarios. We observed that harder scenarios, on average, caused larger changes in the mental model of participants. However, this alone was insufficient to mitigate the observed errors. Furthermore, the consistent error patterns across different participant groups present an opportunity to enhance the current framework of narration and presentation of explanations, benefiting all participants.

\section{Limitations and Future Work}

The experiment provides evidence for the usefulness of global explanations in health informatics. Identifying cognitive styles that lead to particular explanation preferences and errors in comprehension, is pivotal to applying global explanations in real-life applications. The current experiment has been carried out on a small dataset. Evaluating these findings on a larger data set with more data points and larger features will be undertaken in future studies. We recognise that regression models are commonly used in risk prediction. Expanding the scope of the narrative global explanation within the context of regression and assessing its comparative utility against the local explanation will enable the integration of our findings into established risk predictive tools.

Further, the evaluation in this study was crowdsourced and hence the participants are not representative of real-life patients. Most of the participants fall in the age category that does not have a risk of heart disease as predicted by the model. This may have biased their rating. We aim to rectify this by conducting the evaluation on a representative patient population, which would also require addressing ethical concerns.

The current study has not focussed on generating effective global explanations. The use of semifactuals has not addressed the mismatch with the user's mental models. Further, the presentation of Explanation
features is seen to have introduced errors. Effective communication and presentation techniques would be vital in reducing errors. Though we have used a linguistic representation of probability and confidence, the evaluations in this regard remain undone. For risk communication at scale, this is a crucial component. Further research is warranted to delve deeper into these aspects and refine the design and implementation of explanation systems.

\section*{Acknowledgement}
We would like to thank Dr. Sameen Maruf and Prof. Ingrid Zukerman for generously sharing their expertise in utilizing the dataset, continuous support and valuable feedback in designing the experiment. We thank Nikolay Babakov and Prof. Alberto José Bugarín Diz for their feedback throughout the development of this research. We also thank the anonymous reviewers for their feedback which has significantly improved this work. A. Sivaprasad is ESR in the NL4XAI project which has received funding from the European Union’s Horizon 2020 research and innovation programme under the Marie Skłodowska-Curie Grant Agreement No. 860621.

\bibliographystyle{splncs04}
\bibliography{ecai_bibliography}

\appendix 
\appendix

\section{Construction and Selection of DT} 
\label{appendix:ConstDT}

\begin{table}[!h]
\caption{Category definitions for Data preprocessing.}
\label{tab:catDefinition}
\centering
\small
\begin{tabular}{p{3.7cm}|p{8.3cm}}
\toprule
\textbf{Feature} & \textbf{Categories} \\
\midrule
\midrule
Smoking & Non-smoker, light smoker: (less than 10), moderate smoker - (10 to 19)/day, heavy smoker- (20 or over)/day) \\
\hline
BMI &Underweight (less than 18.5), Healthy - (18.5 to 24.9), Overweight - (25 to 29.9), Obese - (30 or over)\\

\hline
Cholesterol & Normal: $\le$5, High: above 5 \\
\hline
Cholestrol HDL ratio & Normal: $\le$6, High: above 6\\
\hline
Triglycerides & Fasting - normal (0 to 1.7), Non Fasting - normal (1.7 to 2.3), High (2.3 to 10) \\
\hline
HDL & Normal: $\le$ 1, High: above 1 \\
\hline
Systolic Blood Pressure & Low: (0 to 90), Normal: (90 to 120), Elevated: (120 to  140), High: (140 to 250)\\
\hline
Diastolic Blood pressure &Low: (0 to 60), Normal: (60 to 80), Elevated: (80 to 90) , High: (90 to 150)\\
\bottomrule
\end{tabular}
\end{table}

\begin{figure}[hbt!]
    \centering
    \includegraphics[width=0.7\linewidth]{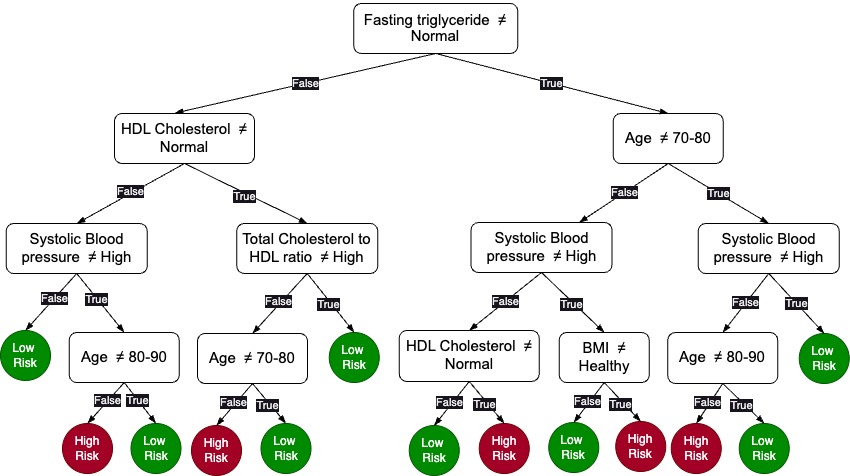}
    \caption{Depth 5 Decision tree generated on 2134 datapoints. Training accuracy = 90.9\% , test accuracy on 534 records = 85\%.}
    \label{fig:DTall}
\end{figure}

\begin{figure}[hbt!]
    \centering
    \includegraphics[width=0.95\linewidth]{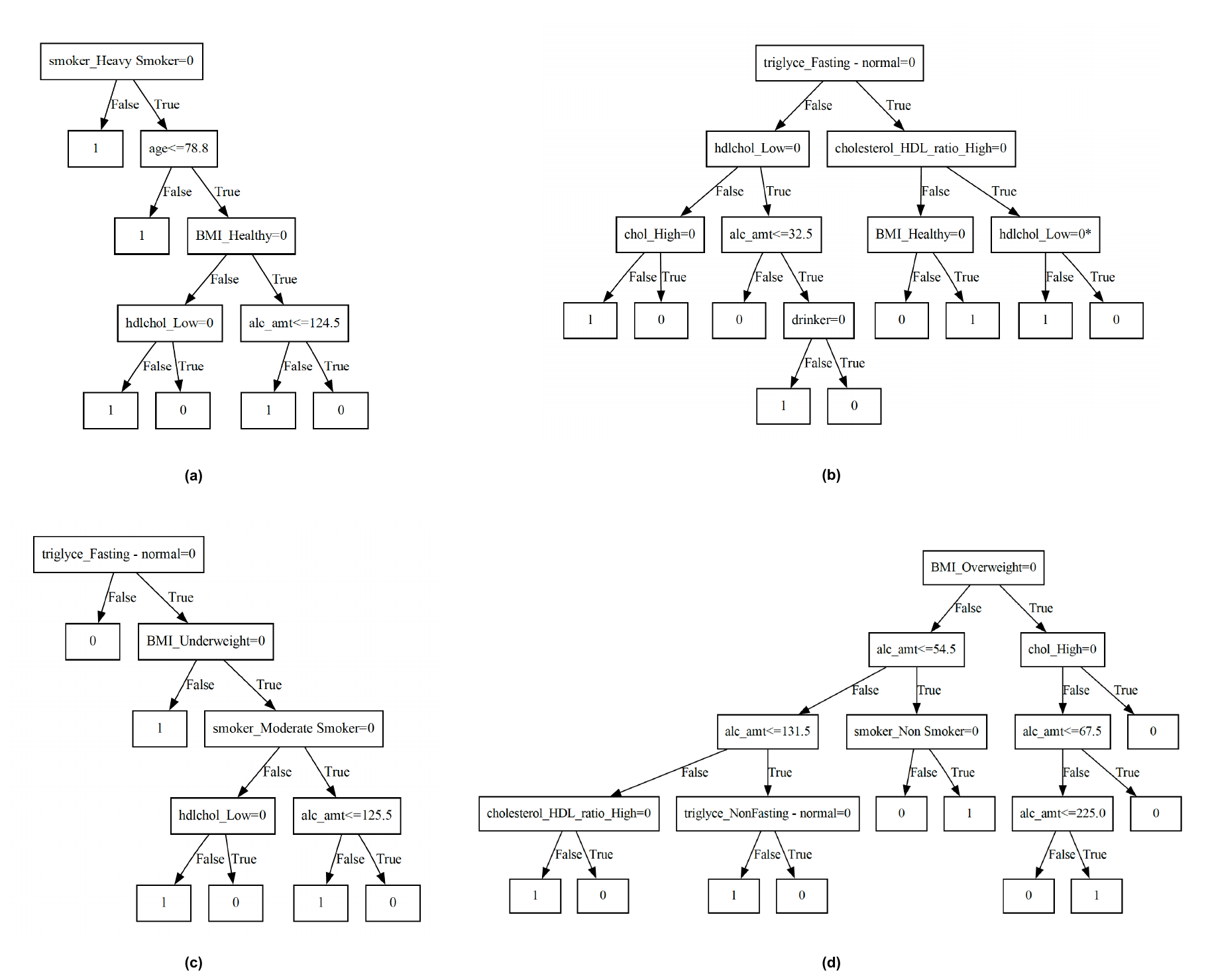}
    \caption{DTs for different scenarios. (a) Local easy scenario: Decision tree generated on 116 data points. Training accuracy = 78.4\%, (b) Local Hard scenario: Decision tree generated on 163 data points. Training accuracy = 77.3\%, (c) Global easy  scenario: Decision tree generated on 382 data points. Training accuracy = 82.5\% (d) Global Hard scenario: Decision tree generated on 108 data points. Training accuracy = 85.4\%.}
    \label{fig:4scenarios}
\end{figure}

\clearpage
\section{Generating Explanation Narration}\label{appendix:genExpNarration}

Steps in generating narration:
\begin{enumerate}
  \item Filter the rules corresponding to high risk leaves.
  \item Sort the decision rules in order of their leaf node confidence and insert verbal mapping of relative probability.
  \item Reorder the features and place contradictory features at the end preceded by even-if.
  \item combine the features with \emph{and} 
  \item Add header with \emph{age}, \emph{gender}
\end{enumerate}

\begin{figure}[hbt!]
    \centering
    \includegraphics[width=0.4\linewidth]{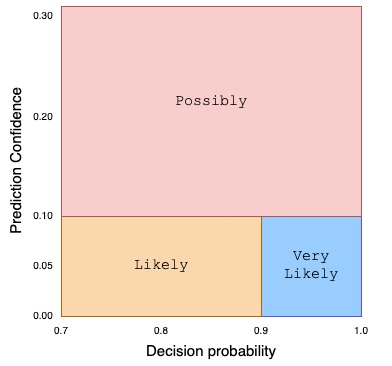}
    \caption{Verbal mapping of relative probabilities.}
    \label{fig:probMapping}
\end{figure}

\begin{figure}
    \centering
    \includegraphics[width=0.6\linewidth]{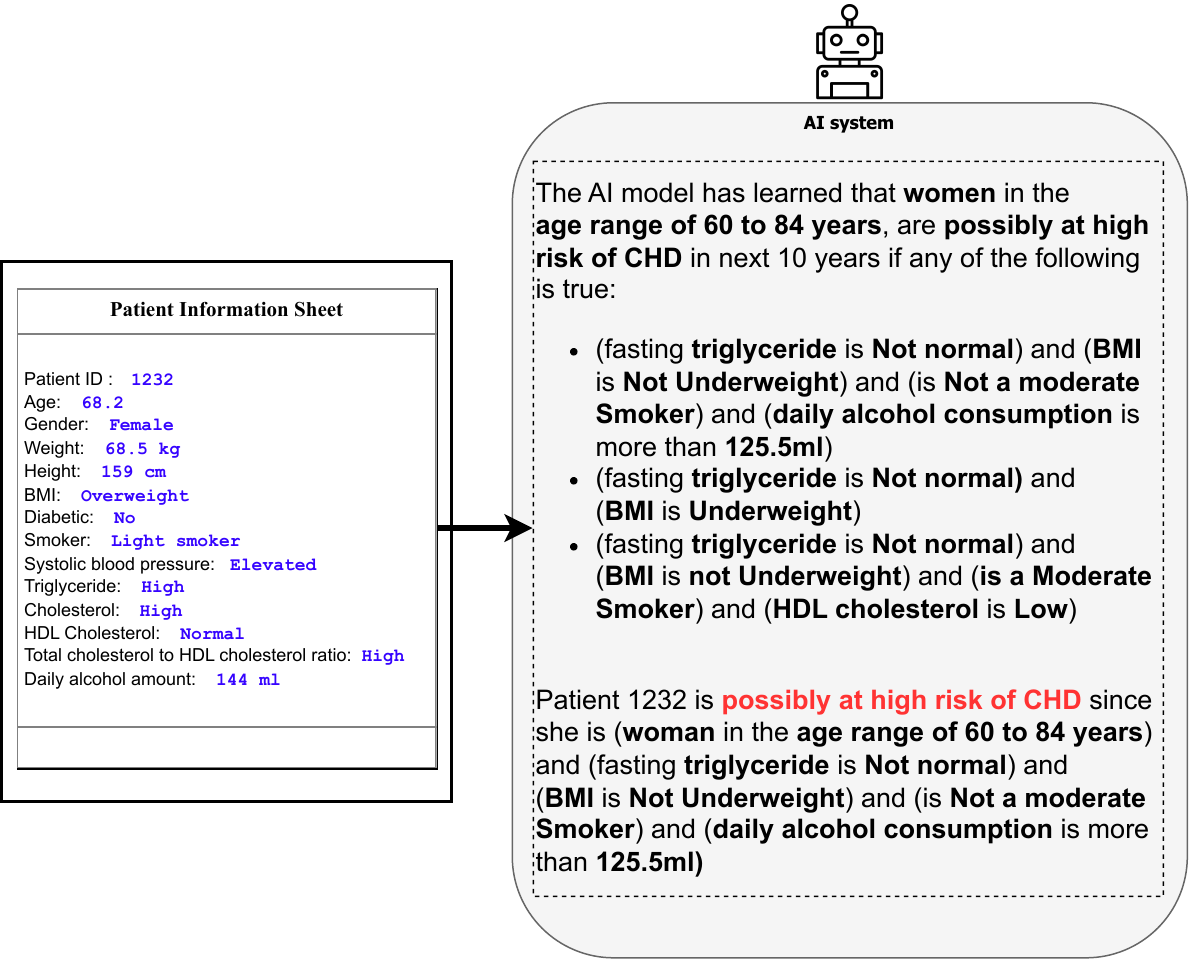}
    \caption{An example for generated global explanation. This model corresponds to Global-easy scenario.}
    \label{fig:globalEasyXplatn}
\end{figure}

\section{User Survey on Prolific}\label{appendix:userSurvey}

\begin{figure}[hbt!]
    \centering
    \includegraphics[width=1\linewidth]{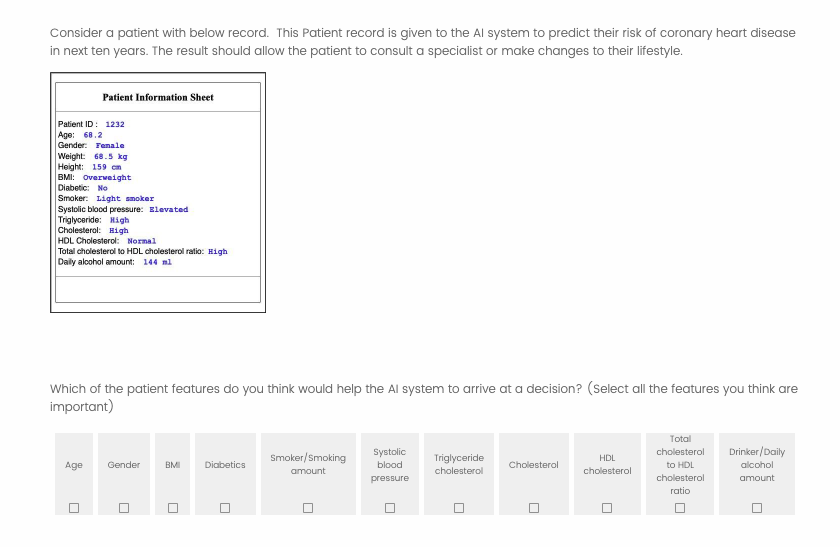}
    \caption{First page of a scenario shown to participants with a patient info. They question captures the participant's mental model of CHD prediction before viewing explanation.}
    \label{fig:surveyScreen1}
\end{figure}

For each scenario, a participant first see the patient information as shown in Figure \ref{fig:surveyScreen1}. The participant is asked to pick all the features they think might be influential in predicting the patient's risk of CHD. This captures the participants mental model regarding CHD prediction before viewing any explanation.

\begin{figure}[hbt!]
    \centering
    \includegraphics[width=0.7\linewidth]{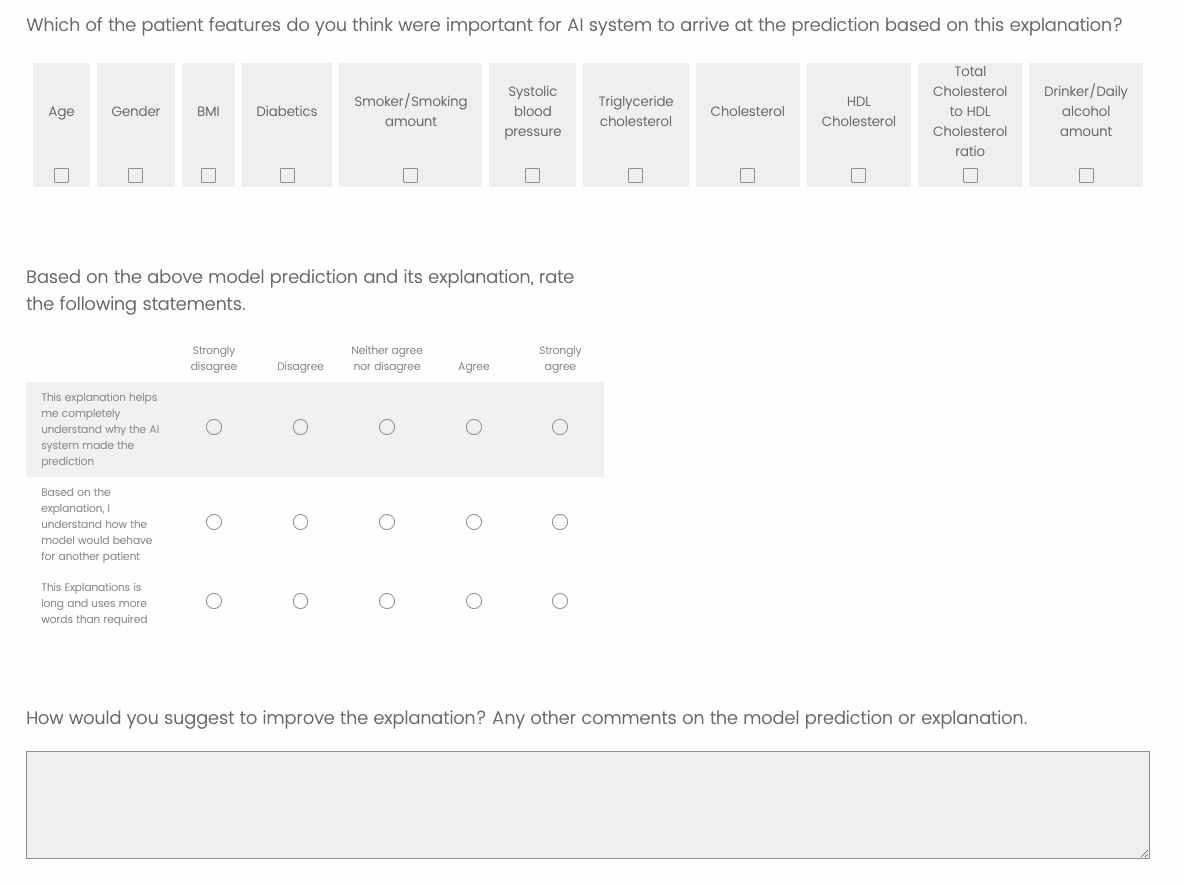}
    \caption{The first question evaluates participant's understanding of the explanation. The Remaining questions capturing their feedback on explanation.}
    \label{fig:surveyScreen2}
\end{figure}

In the next page, a participant is shown the explanation followed by questions to rate the explanation. The are asked to redo the task of picking all the features they think were influential in predicting the patient's risk of CHD as shown in Figure \ref{fig:surveyScreen2}. This captures the participant's understanding of AI's mental model. This is followed by questions to get the users rating based on a 5 point Likert scale. The questions correspond to 3 parameters being measured:
\begin{enumerate}
  \item Completeness : This explanation helps me completely understand why the AI system made the prediction
  \item Understandability : Based on the explanation, I understand how the model would behave for another patient
  \item Verboseness : This explanation is long and uses more words than
require 
\end{enumerate}

\section{Comparison of Local and Global Explanation Ratings}\label{appendix:resultsXTnded}

Results of Wilcoxon test, for combinations of explanation types within participant Groups. After Bonferroni Correction, p-values less than 0.01 are chosen as significant.

\begin{table}[!htbp]
\caption{Significance of difference between different types of explanation for Group 1 rounded to 2 decimal places. Significant p-value are in \textbf{bold}. P-value $\leq{0.001}$ are highlighted with *.}
\label{tab:significanceTestg1}
\centering
\scriptsize
\begin{tabular}{p{4.5cm}p{1.4cm}p{1.4cm}p{1.4cm}p{1.4cm}p{1.4cm}}
\toprule
& CR & UR & VR & CMM & EU \\
\midrule
\midrule
Local vs Global & $\textbf{0.00}^*$ & $\textbf{0.00}^*$ & $\textbf{0.00}^*$ & 0.81 & \textbf{0.00} \\
\midrule
Local easy vs Global easy & \textbf{0.00} & \textbf{0.01} & \textbf{0.00} & 0.93 & \textbf{0.00} \\
\midrule
Local Hard vs Global Hard & 0.05 & 0.05 & 0.03 & 0.62 & 0.07 \\
\midrule
Local easy vs Local Hard & 0.20 & 0.21 & 0.50 & 0.04 & 0.19 \\
\midrule
Global easy vs Global Hard & 0.34 & 0.65 & 0.16 & 0.56 & 0.66 \\
\midrule
Local SHAP vs Local Hard & 0.53 & 0.79 & 0.04 & 0.29 & 0.79 \\
\midrule
Local SHAP vs Local easy & 0.04 & 0.34 & \textbf{0.01} & \textbf{0.01} & 0.18 \\
\midrule
Local SHAP vs Global Hard & 0.21 & 0.03 & \textbf{0.01} & 0.21 & 0.10 \\
\midrule
Local SHAP vs Global easy & 0.03 & 0.07 & \textbf{0.00} & 0.02 & 0.35 \\
\bottomrule
\end{tabular}
\end{table}

\begin{table}[!htbp]
\caption{Significance of difference between different types of explanation for Group 2 rounded to 2 decimal places. Significant p-value are in \textbf{bold}. P-value $\leq{0.001}$ are highlighted with *.}
\label{tab:significanceTestg2}
\centering
\scriptsize
\begin{tabular}{p{4.5cm}p{1.4cm}p{1.4cm}p{1.4cm}p{1.4cm}p{1.4cm}}
\toprule
& CR & UR & VR & CMM & EU \\
\midrule
\midrule
Local vs Global & 0.02 & \textbf{0.01} & 0.13 & 0.89 & $\textbf{0.00}^*$ \\
\midrule
Local easy vs Global easy & 0.22 & 0.08 & 0.10 & 0.38 & $\textbf{0.00}^*$ \\
\midrule
Local Hard vs Global Hard & 0.06 & 0.08 & 0.65 & 0.29 & $\textbf{0.00}^*$ \\
\midrule
Local easy vs Local Hard & 0.40 & 0.60 & 0.92 & 0.02 & 0.19 \\
\midrule
Global easy vs Global Hard & 0.53 & 0.24 & 0.21 & 0.42 & 0.35 \\
\midrule
Local SHAP vs Local Hard & 0.84 & 0.99 & 0.51 & 0.81 & 0.92 \\
\midrule
Local SHAP vs Local easy & 0.27 & 0.71 & 0.97 & 0.05 & 0.58 \\
\midrule
Local SHAP vs Global Hard & \textbf{0.00} & 0.02 & 0.80 & 0.71 & \textbf{0.01} \\
\midrule
Local SHAP vs Global easy & 0.03 & 0.07 & 0.10 & 0.20 & 0.02 \\
\bottomrule
\end{tabular}
\end{table}

\begin{table}[!htbp]
\caption{Significance of difference between different types of explanation for Group 3 rounded to 2 decimal places. Significant p-value are in \textbf{bold}. P-value $\leq{0.001}$ are highlighted with *.}
\label{tab:significanceTestg3}
\centering
\scriptsize
\begin{tabular}{p{4.5cm}p{1.4cm}p{1.4cm}p{1.4cm}p{1.4cm}p{1.4cm}}
\toprule
& CR & UR & VR & CMM & EU \\
\midrule
\midrule
Local vs Global & \textbf{0.01} & 0.05 & $\textbf{0.00}^*$ & 0.17 & $\textbf{0.00}^*$ \\
\midrule
Local easy vs Global Hard & 0.02 & 0.10 & \textbf{0.00} & 0.03 & $\textbf{0.00}^*$ \\
\midrule
Local Hard vs Global Hard & 0.27 & 0.27 & $\textbf{0.00}^*$ & 0.94 & \textbf{0.01} \\
\midrule
Local easy vs Local Hard & 0.60 & 0.62 & 0.77 & 0.14 & \textbf{0.00} \\
\midrule
Global easy vs Global Hard & 0.66 & 0.26 & 0.14 & 0.96 & 0.40 \\
\midrule
Local SHAP vs Local Hard & 0.56 & 0.62 & 0.13 & 0.25 & 0.06 \\
\midrule
Local SHAP vs Local easy & 0.60 & 0.60 & 0.07 & \textbf{0.01} & 0.29 \\
\midrule
Local SHAP vs Global Hard & 0.05 & 0.23 & $\textbf{0.00}^*$ & 0.15 & $\textbf{0.00}^*$ \\
\midrule
Local SHAP vs Global easy & 0.03 & 0.05 & $\textbf{0.00}^*$ & 0.23 & \textbf{0.01} \\
\bottomrule
\end{tabular}
\end{table}

\begin{table}[!htbp]
\caption{Error in selecting patient feature after explanation. Type I error (False Positive) - Wrong selection overall.}
\label{tab:errorType1}
\centering

\begin{tabular}{p{6cm}p{1.5cm}p{1.5cm}p{1.5cm}p{1.5cm}p{1.5cm}}
\toprule
& Local SHAP & Local easy & Local hard & Global easy & Global hard \\
\midrule
\midrule
Age& & & & & \\
\midrule
Gender&3& & & & \\
\midrule
BMI & & &2& & \\
\midrule
Diabetics  &5 &2& &1& \\
\midrule
Cholesterol &5&2&2&8& \\
\midrule
HDL cholesterol & & & & &15\\
\midrule
Triglyceride cholesterol&1& & & & \\
\midrule
Total cholesterol to HDL cholesterol ratio&2&1&1&6& \\
\midrule
Systolic blood pressure&5&1&&2&5\\
\midrule
Smoking/ Smoking amount& & &2& &\\
\midrule
Dinker/ Drinking amount& & & & &2\\
\bottomrule
\end{tabular}
\end{table}

\begin{table}[t]
\caption{Error in selecting patient feature after explanation. Type II error (False Negative) - Missing correct feature.}
\label{tab:errorType2}
\centering
\begin{tabular}{p{6cm}p{1.5cm}p{1.5cm}p{1.5cm}p{1.5cm}p{1.5cm}}
\toprule
& Local SHAP & Local easy & Local hard& Global easy & Global hard\\
\midrule
\midrule
Age& 6&6&1&8&9\\
\midrule
Gender& &8&13&22&23\\
\midrule
BMI &14&1&&1&3\\
\midrule
Diabetics  & & & & & \\
\midrule
Cholesterol & & & & &31\\
\midrule
HDL cholesterol &10& &23&37& \\
\midrule
Triglyceride cholesterol& & &20& 4&35\\
\midrule
Total cholesterol to HDL cholesterol ratio& & & & &11 \\
\midrule
Systolic blood pressure& & & & & \\
\midrule
Smoking/ Smoking amount&4&17& & 10&27\\
\midrule
Dinker/ Drinking amount&12& &9&3&\\
\bottomrule
\end{tabular}
\end{table}
\clearpage

\end{document}